\documentclass[10pt,letterpaper]{article}

\usepackage{spconf}
\usepackage[T1]{fontenc}
\usepackage[utf8]{inputenc}
\usepackage{amsmath,amssymb}
\usepackage{newtxtext,newtxmath}
\usepackage{tipa}
\DeclareUnicodeCharacter{0261}{{\fontfamily{cmr}\selectfont\textipa{g}}}
\usepackage{textcomp}
\usepackage{microtype}
\usepackage{booktabs}
\usepackage{flushend}
\usepackage{needspace}
\usepackage{tabularx}
\usepackage{array}
\usepackage{enumitem}
\usepackage{graphicx}
\usepackage[table]{xcolor}
\usepackage{tikz}
\usepackage{pgfplots}
\usepackage{cite}
\usepackage{xurl}
\usepackage[hidelinks]{hyperref}
\usepackage[nameinlink,noabbrev]{cleveref}
\hypersetup{
  pdftitle={BranchShine-CR: Compact Multilingual IPA Transcription with Self-Conditioned CTC and Consistency Regularization},
  pdfauthor={Nikhil Navas; Sergio Chevtchenko; Talisson Damiao; Saeed Afshar},
  pdfsubject={IPApack++ recognition with reduced-budget development ablations},
  pdfkeywords={IPA transcription, E-Branchformer, self-conditioned CTC, consistency regularization}
}

\graphicspath{{figures/}}

\pgfplotsset{compat=1.18}
\usetikzlibrary{positioning,arrows.meta,calc}

\setlist[itemize]{leftmargin=*,topsep=2pt,itemsep=1pt,parsep=0pt}
\newcolumntype{Y}{>{\raggedright\arraybackslash}X}
\newcommand{\model}{BranchShine-CR}
\newcommand{\ipacer}{\textsc{IPA-CER}}
\newcommand{\pfer}{\textsc{PFER}}

\newcommand{\crcer}{4.47}
\newcommand{\crexact}{41.37}
\newcommand{\crpfer}{2.09}
\newcommand{\crmacro}{10.18}
\newcommand{\crvsoriginal}{37.2}
\newcommand{\crvsnemo}{48.4}
\newcommand{\crvszipa}{22.3}
\newcommand{\crsmaller}{23.9}

\definecolor{branchblue}{RGB}{31,78,121}
\definecolor{branchlight}{RGB}{225,235,245}
\definecolor{softgray}{gray}{0.55}
\definecolor{palegray}{gray}{0.92}

\title{BranchShine-CR: Compact Multilingual IPA Transcription with Self-Conditioned CTC and Consistency Regularization}
\name{
Nikhil Navas$^{1}$ \quad Sergio Chevtchenko$^{1}$ \quad Talisson Damiao$^{2}$ \quad Saeed Afshar$^{1}$}
\address{
$^{1}$International Centre for Neuromorphic Systems, Western Sydney University\\
$^{2}$Neurabuild
}

\begin{document}
\ninept
\maketitle

\begin{abstract}
We introduce \model{}, a 25M-parameter model for multilingual transcription into the International Phonetic Alphabet (IPA). It combines log-mel features, a rotary-position E-Branchformer encoder, intermediate self-conditioned connectionist temporal classification (CTC), and consistency regularization across augmented views. On 16,646 shared IPApack++ test utterances, it achieves \crcer\% IPA character error rate, a \crvszipa\% relative reduction from ZIPA-CTC-NS, with approximately one-twelfth as many parameters while being trained from scratch. \model{} also outperforms a similarly sized NeMo Conformer baseline across all 41 dataset language labels. Ablation studies indicate the individual components synergetically acting in model performance contribution. These findings support compact IPA recognition capabilities under limited compute budget, for applications in low-resource on-device pronunciation assessment.
\end{abstract}

\section{Introduction}

Automatic speech recognition typically produces orthographic text. For language documentation, pronunciation analysis, and cross-lingual speech research, a transcription of the sounds a speaker produces can be more useful. Direct transcription into the International Phonetic Alphabet (IPA) offers a shared representation across languages \cite{lee2024learner}.

Recent multilingual phone recognizers benefit from large training resources and pretrained speech encoders \cite{zhu2025zipa,bharadwaj2026phoneticxeus}. We investigate how much recognition accuracy a compact model trained from scratch can provide. Our original BranchShine model \cite{navas2026branchshine} uses a learned raw-waveform front end and a RoPE E-Branchformer encoder. Here, we retain that model as a reference and introduce \model{}, which uses log-mel features, a smaller encoder, intermediate CTC conditioning, and a consistency-regularized training objective.

We compare six systems on identical IPApack++ test utterances and references. \model{} combines a smaller parameter count with lower IPA character error (\ipacer{}), while ZIPA-CTC-NS retains higher exact match. Language and edit analyses examine the gains across groups and error types. A separate ablation study is presented to examine the contribution of the individual components.

\section{Related Work}

\subsection{Multilingual phone recognition}
AlloVera and Allosaurus established shared resources and models for phone recognition across languages \cite{mortensen2020allovera,li2020allosaurus}. Subsequent work explored compositional phone representations, language-specific inventories, and articulatory supervision \cite{li2021hierarchical,li2022phone,glocker2023allophant}. Wav2Vec2Phoneme combined multilingual representations with articulatory mappings \cite{xu2022simple}; MultiIPA investigated direct IPA transcription and cleaner multilingual training data \cite{taguchi2023multiipa}. ZIPA uses IPApack++ for multilingual training \cite{zhu2025zipa}, while POWSM jointly addresses phone and orthographic recognition and conversion between phonetic and written forms \cite{li2025powsm}. PhoneticXEUS combines the XEUS multilingual encoder with self-conditioned CTC \cite{chen2024xeus,bharadwaj2026phoneticxeus}. These systems provide reference points for a compact recognizer trained from scratch and are used as experimental baseline in the present work.

\subsection{Encoder design and supervision}
Branchformer models global and local acoustic context through parallel self-attention and convolutional-gating branches, while E-Branchformer strengthens the fusion of these representations through enhanced branch merging \cite{peng2022branchformer,kim2022ebranchformer}. Rotary position embeddings (RoPE) provide positional information within self-attention \cite{su2021roformer}, and recent compact recognizers such as Moonshine further motivate efficient speech encoder design \cite{jeffries2024moonshine}. Intermediate CTC introduces auxiliary supervision at internal encoder layers \cite{lee2021interctc}, whereas self-conditioned CTC additionally feeds intermediate posterior distributions back into the hidden representation to condition subsequent layers \cite{nozaki2021selfctc}. Consistency regularization is related to R-Drop, which encourages agreement between stochastic dropout predictions \cite{liang2021rdrop}. Our objective instead forms two stochastic views of the same speed-perturbed waveform, using independently sampled masking and dropout, and penalizes disagreement between their final CTC posteriors with symmetric KL divergence, stopping gradients through the target distribution in each direction. These mechanisms are individually established and our contribution is their integration into a compact, from-scratch multilingual IPA recognizer and its empirical evaluation\cite{yao_cr-ctc_2025}. PanPhon's articulatory feature representation also provides a complementary diagnostic to exact IPA character identity \cite{mortensen2016panphon}.

\section{Model and Training}
\label{sec:model}

\subsection{Acoustic encoder and conditioning}
\model{} accepts mono 16~kHz audio and computes 80-bin log-mel power features using a 25~ms Hann window, a 512-point FFT, and a 10~ms hop. Each frequency bin is normalized over valid frames of the utterance. Two $3\times3$ stride-2 convolutions, with 32 and 64 channels, reduce time and frequency resolution. The result is then projected to 256 dimensions.

\begin{figure}[t]
\centering
\includegraphics[width=0.9\columnwidth]{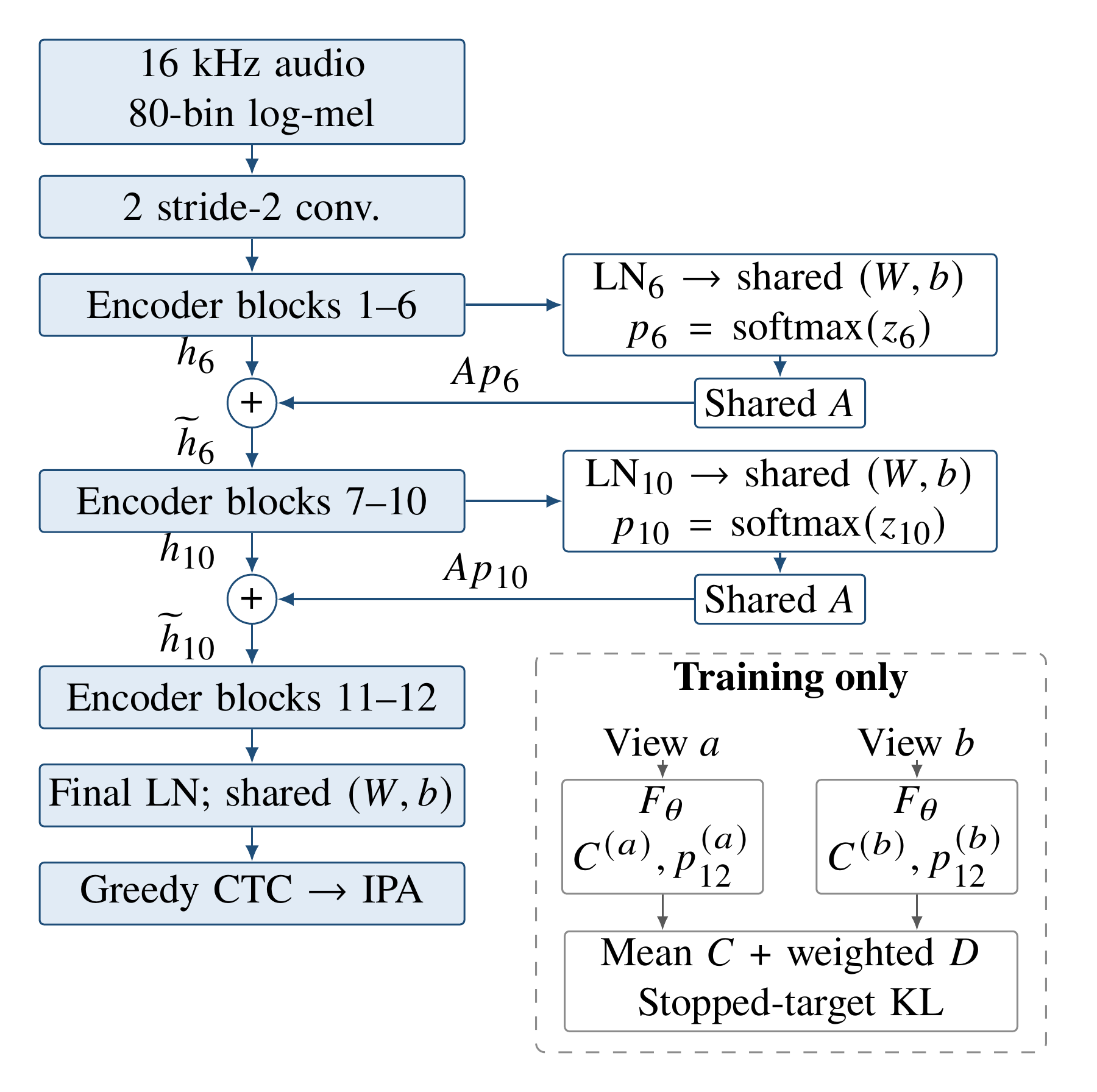}
\caption{\model{} architecture. Shared $(W,b)$ produce the logits; shared $A$ projects posteriors into the $+$ nodes during training and inference. The inset uses two views of network $F_\theta$: $C$ combines final and auxiliary CTC; $D$ is symmetric stopped-target KL.}
\label{fig:architecture}
\end{figure}

The encoder contains 12 RoPE E-Branchformer blocks. Each block has two half-scaled feed-forward residual modules of width 1,024 around parallel four-head attention and convolutional gating branches. The gating branch uses two 768-channel halves and a 1-D  depthwise convolution of width 31 along the time dimension. Concatenated attention and local features pass through a depthwise residual merge of kernel size 31 and a projection back to 256 dimensions. \Cref{fig:architecture,tab:model-config} summarize the signal path and contrast it with the original BranchShine model \cite{navas2026branchshine}.

\begin{table}[t]
\centering
\caption{Configurations of the two BranchShine systems.}
\label{tab:model-config}
\small
\begin{tabularx}{\columnwidth}{@{}Yrr@{}}
\toprule
Component & Original & CR \\
\midrule
Front end & Raw waveform & 80-bin log-mel \\
Encoder blocks & 19 & 12 \\
Hidden size / heads & 288 / 8 & 256 / 4 \\
Feed-forward width & 480 & 1,024 \\
Intermediate CTC layers & None & 6, 10 \\
Vocabulary size & 112 & 112 \\
Parameters & 33,381,712 & 25,393,904 \\
\bottomrule
\end{tabularx}
\end{table}
After blocks $\ell\in\{6,10\}$, hidden states $h_\ell$ undergo
layer-specific normalization $\mathrm{LN}_\ell$ and a shared
CTC projection $(W,b)$:
\begin{equation}
z_\ell=W\,\mathrm{LN}_\ell(h_\ell)+b,\qquad
\widetilde h_\ell=h_\ell+A\,\mathrm{softmax}(z_\ell).
\end{equation}
Here $z_\ell$ are vocabulary logits; shared $A$ maps their
posteriors to the hidden dimension, and $\widetilde h_\ell$
feeds the next block. Operations are framewise. A final
normalization and the same $(W,b)$ produce logits at block 12.
The 112-symbol vocabulary includes the CTC blank and a space
token. Inference retains prediction feedback and uses one
unaugmented view with greedy CTC decoding, without a language
model or beam search.

\subsection{Two-view training objective}
CTC maps acoustic sequences to target strings without
frame-level alignments \cite{graves2006ctc}. Two independently
masked views $a,b$ of the same speed-perturbed waveform pass
through network $F_\theta$ with shared parameters $\theta$.
For view $v\in\{a,b\}$ and target strings $y$, the supervised
objective is
\begin{equation}
C^{(v)}=0.7\,\mathcal L_{\mathrm{CTC}}(z_{12}^{(v)},y)
+0.15\sum_{\ell\in\{6,10\}}
\mathcal L_{\mathrm{CTC}}(z_\ell^{(v)},y),
\end{equation}
where $z_\ell^{(v)}$ denotes layer-$\ell$ logits for view $v$.
For a batch of $B$ utterances, let $p_{jt}^{(v)}$ be the final
vocabulary posterior at frame $t$ of utterance $j$, with $T_j$
valid frames. The consistency loss is
\begin{equation}
\begin{split}
D=\frac{1}{2B}\sum_{j=1}^{B}\sum_{t=1}^{T_j}\big[&
\mathrm{KL}(\mathrm{sg}(p_{jt}^{(a)})\Vert p_{jt}^{(b)})\\
+&\mathrm{KL}(\mathrm{sg}(p_{jt}^{(b)})\Vert p_{jt}^{(a)})\big],
\end{split}
\end{equation}
where $\mathrm{KL}$ denotes Kullback--Leibler divergence
and $\mathrm{sg}$ stops the target gradient. Padding frames
are excluded. At optimizer step $s$, the full objective is
\begin{equation}
\mathcal L=\frac{C^{(a)}+C^{(b)}}{2}
+0.2\min(1,s/2000)\,D.
\end{equation}
The consistency weight ramps to 0.2 over 2,000 updates.
CTC losses use an utterance mean without target-length
normalization. Consistency is summed over valid frames
and averaged over utterances and both directions.

Training uses speed factors $\{0.9,1.0,1.1\}$, restricted to those preserving CTC feasibility. Each view uses dropout 0.1 and independently sampled SpecAugment \cite{park2019specaugment} with application probability 0.9, two frequency masks of width up to 27, and adaptive time masks with a 37.5\% total-width budget and at most 25 masks. Timing is unchanged between the two views.

We train from randomly initialized weights with AdamW optimizer  ($\beta_1=0.9$, $\beta_2=0.999$), peak learning rate $5\times10^{-4}$, 4,000-step warmup, and cosine decay to 5\% of the peak rate. Weight decay is 0.01 for parameters of dimension at least two, while other parameters receive no decay. Gradient norm is clipped at 1.0. Three microbatches are accumulated per optimizer step, each capped at 256~s of audio or 128 utterances. The run uses BF16, activation checkpointing, and an RTX PRO 5000 Blackwell GPU. It completes 350,000 updates without early stopping in 4.7 days and the exported model is selected at step 346,000 by development character-token error including spaces.

\section{Experimental Setup}
\label{sec:protocol}

\begin{table*}[t]
\centering
\caption{IPApack++ results on the same 16,646 test utterances with identical references and normalization. Lower is better for \ipacer{} and \pfer{}; higher is better for exact match. Training regimes differ.}
\label{tab:leaderboard}
\small
\begin{tabularx}{\textwidth}{@{}Yrrrr@{}}
\toprule
Model & Params (M) & \ipacer{} (\%) & Exact match (\%) & \pfer{} (\%) \\
\midrule
\rowcolor{branchlight}
\model{} & 25.39 & \textbf{4.47} & 41.37 & \textbf{2.09} \\
ZIPA-CTC-NS & 299.97 & 5.76 & \textbf{45.15} & 2.14 \\
ZIPA-CTC & 299.97 & 6.51 & 38.43 & 2.48 \\
Original BranchShine & 33.38 & 7.13 & 26.03 & 3.07 \\
NeMo Conformer-CTC Medium & 30.53 & 8.67 & 19.55 & 3.84 \\
PhoneticXEUS & 575.00 & 9.76 & 20.19 & 3.20 \\
\bottomrule
\end{tabularx}

\end{table*}

\subsection{Data and comparison systems}
We use the training partition \#3 from IPApack++ for the main recognition experiments. Our canonical partition contains 1,632,681 training utterances and 16,661 utterances in each of the development and test splits, including English. Due to the specific CTC-feasibility requirement of each model, different data-preparation strategies are adopted. Hence, \model{} retains 1,631,436 training utterances, original BranchShine 1,632,596, and NeMo 1,631,217. Similarly, their development evaluation counts differ slightly: 16,661, 16,659, and 16,639, respectively. 

All six systems are evaluated on a shared set of 16,646 test utterances from this partition, totaling 25.6 hours, with 818,485 normalized reference characters. The shared set has 41 stored language labels. The five largest labels account for 67.9\% of utterances. We therefore report corpus-level error, the unweighted mean of label-level error rates, and a sensitivity check on the dataset.

Original BranchShine uses three waveform convolutions (kernels 127, 7, 3; strides 64, 3, 2) and 19 RoPE E-Branchformer blocks. We retain its development-selected step-1,801,000 checkpoint. NeMo Conformer-CTC Medium uses 80-bin log-mel features, 18 Conformer blocks, and 30.53M parameters \cite{gulati2020conformer}, its completed 350,000-step run selects step 346,000. Both use greedy CTC decoding. All three compact systems are trained from scratch on the canonical split, with recommended filters, objectives, schedules, and training budgets. 

The remaining three baseline rows (ZIPA-CTC-NS, ZIPA-CTC, and PhoneticXEUS) use the existing saved predictions. ZIPA and PhoneticXEUS use different IPApack++ training splits, and as per their respective training regimen, are trained on the entire ipapack corpus, which includes splits 1 to 4, and an overlap with the present test set has not been ruled out, making this comparison potentially skewed against our proposed models. 

\subsection{Transcription metrics}
For reference $y_i$ and prediction $\hat y_i$, normalization $N$ applies Unicode NFC, maps ASCII ``g'' to IPA script-g ({\fontfamily{cmr}\selectfont\textipa{g}}), and removes all whitespace. We define
\begin{equation}
 \mathrm{IPA\mbox{-}CER}=100\,\frac{\sum_i\mathrm{ED}(N(y_i),N(\hat y_i))}{\sum_i|N(y_i)|},
\end{equation}
where $\mathrm{ED}$ is Unicode-character Levenshtein distance. Exact match is the percentage of utterances with identical normalized strings. These are character-level metrics, the training log's ``PER'' label also denotes character-token error and should not be interpreted as segmented phone error.

Phonetic feature error rate (\pfer{}) is the summed PanPhon feature-edit cost divided by the number of reference phones, multiplied by 100. We use the preserved PanPhon 0.22.2 scorer and its additional NFD normalization for segmentation. Insertion and deletion costs average feature costs of 0.5 for unspecified features and 1 otherwise. Consequently, substitution costs average half the absolute feature-vector differences. The shared denominator is 778,593 parsed reference phones. Unrecognized material is skipped by the parser, so we retain parse-warning coverage and use \pfer{} as a supplementary diagnostic.

\begin{figure*}[t]
\centering
\includegraphics[width=\textwidth]{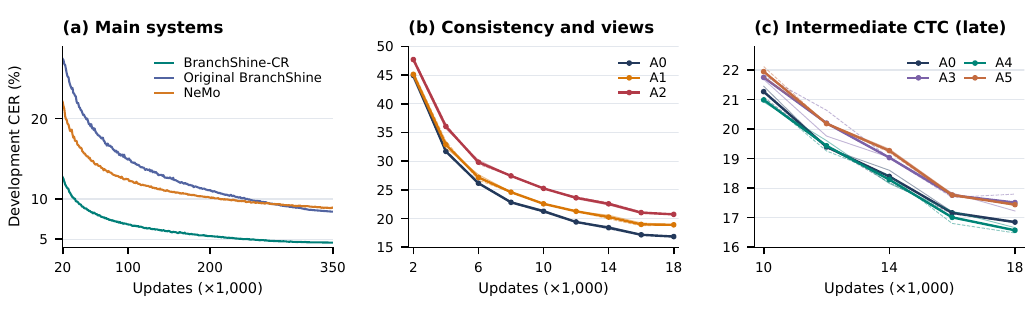}
\caption{Development CER trajectories, without smoothing. (a) Main systems over 20k--350k updates, using native scores including spaces and 16,661/16,659/16,639 development utterances for CR/original/NeMo. (b,c) Separate 18k ablation study on 6,069 utterances: consistency and views, and a 10k--18k detail of intermediate CTC. Thick lines show two-seed means and thin solid/dashed lines show seeds 17/29. Variant definitions are in \Cref{tab:cr-ablations}.}
\label{fig:training_curves}
\end{figure*}

\section{Results}
\label{sec:results}

\subsection{Matched multilingual recognition}
\model{} makes 36,607 character edits, yielding \crcer\% \ipacer{} (\Cref{tab:leaderboard}): a \crvszipa\% relative reduction from ZIPA-CTC-NS with approximately one-twelfth as many parameters. ZIPA-CTC-NS retains higher exact match, 45.15\% versus \crexact\%. CR reduces character error by \crvsoriginal\% relative to original BranchShine while using \crsmaller\% fewer parameters, and by \crvsnemo\% relative to NeMo. Its \crpfer\% \pfer{} is also lowest among the compared systems.

\Cref{fig:training_curves}(a) shows lower logged development CER for CR throughout the displayed range. These native scores retain spaces and model-specific development coverage, distinct from the normalized matched-test metric.

The full 16,661-utterance CR test export yields 4.51\% \ipacer{} and 41.34\% normalized exact match. Its native evaluator reports 4.57\% character error and 36.81\% exact match including spaces. These populations and normalizations are kept separate from the matched main table.

\subsection{Language labels and edit types}
CR has lower \ipacer{} than original BranchShine and NeMo on all 41 language labels, versus 31 labels for ZIPA-CTC-NS and 38 for PhoneticXEUS. Its unweighted mean of label-level error rates is \crmacro\%, compared with 14.36\% for original BranchShine and 17.61\% for NeMo. 

Excluding both Tamil labels retains 15,499 utterances. Corpus \ipacer{} becomes 4.59\% for CR, 5.90\% for ZIPA-CTC-NS, 7.39\% for original BranchShine, and 8.94\% for NeMo, preserving their ordering. The improvement is therefore not confined to the two Tamil labels or the largest labels.

\begin{figure}[t]
\centering
\includegraphics[width=\columnwidth]{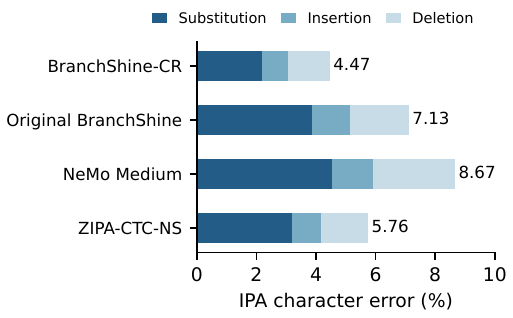}
\caption{Character-error decomposition on the matched test set. CR reduces all three edit types relative to the displayed comparison systems. Bars sum to corpus \ipacer{}.}
\label{fig:error_mix}
\end{figure}

CR makes 17,805 substitutions, 7,253 insertions, and 11,549 deletions, each lower than original BranchShine, NeMo, and ZIPA-CTC-NS under identical deterministic alignment (\Cref{fig:error_mix}). It is worth noting that PanPhon flags unrecognized material in 868 CR reference/prediction pairs (5.21\%) versus 874 (5.25\%) for ZIPA-CTC-NS and their recognized segments remain in the feature score.

\subsection{Ablation study}
\label{sec:ablations}
We evaluate nine configurations for 18,000 updates (\Cref{tab:cr-ablations}). A separate preparation uses 117 symbols, 100,089 training utterances (208.8~h; 142 labels), and 6,069 development utterances (22 labels). Subsets are sampled proportionally within language labels. Scratch training uses 1,500 warmup updates, then a constant learning rate of $5\times10^{-4}$. Common parameters have matched initialization within each seed. Augmentation masks are pinned by update, microbatch, and view.

\begin{table}[t]
\centering
\caption{Ablation studies: final-step development CER (\%) at 18,000 updates. $\Delta$ is the mean paired difference from A0 in percentage points where positive is worse. All variants use seeds 17 and 29, the dash denotes the baseline comparison.}
\label{tab:cr-ablations}
\small
\begin{tabularx}{\columnwidth}{@{}Yrrrr@{}}
\toprule
Variant & Seed 17 & Seed 29 & Mean & $\Delta$ (pp) \\
\midrule
A0: Full CR & 16.87 & 16.83 & 16.85 & --- \\
A1: No consistency, two views & 18.92 & 18.88 & 18.90 & +2.05 \\
A2: Single view, no consistency & 20.70 & 20.74 & 20.72 & +3.87 \\
A3: No auxiliary supervision & 17.23 & 17.80 & 17.51 & +0.66 \\
\rowcolor{branchlight}
A4: No prediction feedback & 16.66 & 16.48 & 16.57 & -0.28 \\
A5: No intermediate CTC & 17.41 & 17.48 & 17.44 & +0.59 \\
A6: No enhanced merge & 17.71 & 17.73 & 17.72 & +0.87 \\
A7: No RoPE & 17.00 & 16.83 & 16.91 & +0.06 \\
A8: Attention only & 18.94 & 18.36 & 18.65 & +1.80 \\
\bottomrule
\end{tabularx}

\end{table}
The table reports final-step weights for 18 runs (A0--A8, seeds 17 and 29). Native Unicode-character CER uses identical references totaling 303,024 characters, without text normalization. References contain no whitespace. The main normalization changes each score by less than 0.01 percentage points. These development scores are separate from the main test leaderboard.

A1 retains two views while removing consistency, A2 also removes the second view. A3 removes auxiliary supervision, A4 removes posterior feedback, and A5 removes both. The main CTC coefficient remains 0.7 when auxiliary supervision is removed, avoiding an increase in main-loss weight. A6 removes the enhanced merge convolution, A7 removes RoPE. A8 removes the local branch, retaining attention only.

Removing consistency increases mean CER by 2.05 percentage points. Removing the second view adds 1.82 points. \Cref{fig:training_curves}(b,c) show the corresponding trajectories and intermediate-CTC comparisons. Removing auxiliary supervision or enhanced merging costs 0.66 or 0.87 points. Removing feedback improves CER by 0.21 and 0.34 points across seeds, removing RoPE has little effect. Removing the local branch costs 1.80 points but reduces parameters from 25.40M to 17.09M, so capacity is not matched.

\section{Conclusion}
Trained from scratch, \model{} achieves \crcer\% \ipacer{} with 25.39M parameters on the matched IPApack++ test set: a \crvszipa\% relative reduction from ZIPA-CTC-NS and \crvsoriginal\% from original BranchShine. ZIPA-CTC-NS retains higher exact match. Reduced-budget ablations favor consistency regularization, while feedback removal slightly improves accuracy and RoPE has little effect. These component findings remain conditional on the ablation data and budget. 

\section{Generative AI use disclosure}
Generative AI tools were used for language refinement and limited coding assistance. The authors have reviewed all AI-assisted material before incorporating it into the manuscript or codebase.

\section{Compliance with Ethical Standards}
\label{sec:ethics}
This research study was conducted retrospectively using human subject data made available in open access by authors of the ZIPA architecture \cite{zhu2025zipa}. Ethical approval was not required as confirmed by the license attached with the open access data.

\clearpage
\section{Conflict of Interest Disclosure}
\label{sec:acknowledgments}
The authors have no relevant financial or nonfinancial interests to disclose..

\bibliographystyle{IEEEbib}
\bibliography{refs}

\end{document}